\documentclass[sigconf]{acmart}
\usepackage{multirow}
\usepackage{colortbl} 

\AtBeginDocument{%
  }

\copyrightyear{2026}
\acmYear{2026}
\setcopyright{cc}
\setcctype{by}
\acmConference[MM '26] {Proceedings of the 34th ACM International Conference on Multimedia}{November 10--14, 2026}{Rio de Janeiro, Brazil.}
\acmBooktitle{Proceedings of the 34th ACM International Conference on Multimedia (MM '26), November 10--14, 2026, Rio de Janeiro, Brazil}
\acmISBN{979-8-4007-2213-4/2026/11}
\acmDOI{10.1145/3767308.3834967}

\begin{document}

\title{OmniAD: Intrinsic Anomaly Detection and Reasoning via Semantic Anomaly Encoding}

\author{Shifang Zhao}
\affiliation{%
    \department{Institute of Information Science, Beijing Jiaotong University}
  \city{Beijing}
  \country{China}}
\email{shifang.zhao@bjtu.edu.cn}

\author{Yiheng Lin}
\affiliation{%
    \department{Institute of Information Science, Beijing Jiaotong University}
  \city{Beijing}
  \country{China}}
\email{23120298@bjtu.edu.cn}

\author{Yunpeng Huan}
\affiliation{%
    \department{Institute of Information Science, Beijing Jiaotong University}
  \city{Beijing}
  \country{China}}
\email{peng0v0zz@gmail.com}

\author{Weizhe Liu}
\affiliation{%
    \department{Institute of Information Science, Beijing Jiaotong University}
  \city{Beijing}
  \country{China}}
\email{weizhe.liu@bjtu.edu.cn}

\author{Lu Han}
\affiliation{%
    \department{Institute of Acoustics, Chinese Academy of Sciences}
  \city{Beijing}
  \country{China}}
\email{hanlu2023@mail.ioa.ac.cn}

\author{Yao Zhao}
\affiliation{%
    \department{Institute of Information Science, Beijing Jiaotong University}
  \city{Beijing}
  \country{China}}
\email{yzhao@bjtu.edu.cn}

\author{Yunchao Wei}
\correspondingauthor
\affiliation{%
  \department{Institute of Information Science, Beijing Jiaotong University}
  \department{Visual Intelligence + X International Joint Laboratory}
  \department{BAAI}
  \city{Beijing}
  \country{China}}
\email{wychao1987@gmail.com}

\renewcommand{\shortauthors}{Shifang Zhao et al.}

\begin{teaserfigure}
\begin{center}
\centering
\vspace{-1.5em}
\includegraphics[width=1\textwidth]{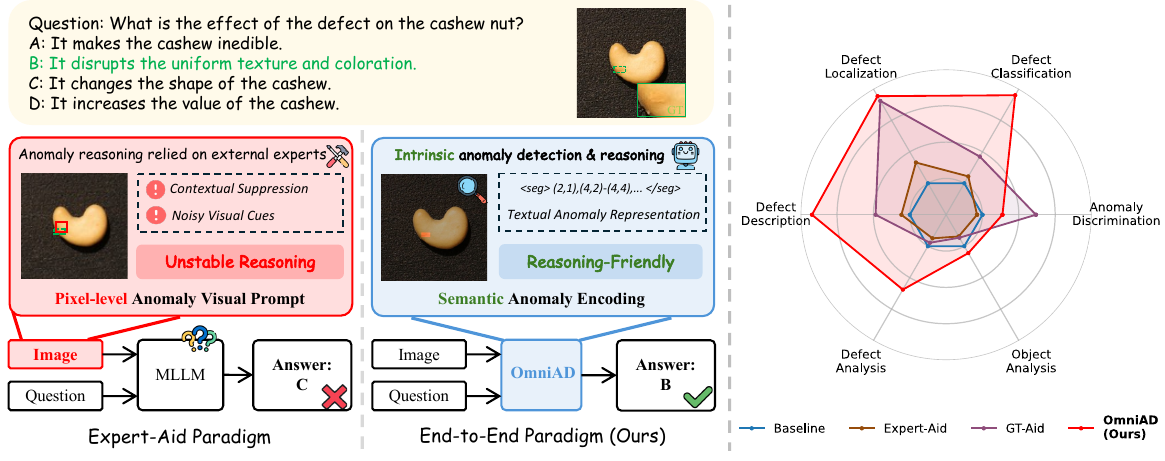}
\end{center}
\vspace{-1.5em}
\caption{\textbf{Overview of the OmniAD framework.} OmniAD is an end-to-end framework that integrates both anomaly detection and anomaly understanding. (Left) A conceptual comparison of the Expert-Aid paradigm, and the intrinsic OmniAD framework. (Right) Quantitative results on the MMAD dataset~\cite{jiang2024mmad}, illustrating the superior performance of OmniAD, where the Expert-Aid paradigm leads to a performance degradation.}
\label{fig:teaser}
\end{teaserfigure}

\begin{abstract}
  A clear and well-documented \LaTeX\ document is presented as an
  article formatted for publication by ACM in a conference proceedings
  or journal publication. Based on the ``acmart'' document class, this
  article presents and explains many of the common variations, as well
  as many of the formatting elements an author may use in the
  preparation of the documentation of their work.
\end{abstract}





\begin{abstract}
Industrial anomaly analysis requires more than anomaly presence prediction: the model must localize subtle defects and explain their causes and implications. Existing MLLM-based solutions often rely on an external anomaly detector to provide visual cues, yet this expert-aid paradigm is brittle because the cues are threshold-sensitive, potentially noisy, and can suppress broader contextual reasoning. We propose OmniAD, a end-to-end framework that intrinsically couple anomaly detection and anomaly understanding within a single multimodal reasoning process. OmniAD first performs visual reasoning through Semantic Anomaly Encoding, which converts anomaly regions into a compact textual representation and removes the need for thresholded heatmaps. It then performs Visual Guided Textual Reasoning, where the internally generated anomaly evidence guides high-level analysis without introducing an external attention bias. To improve learning under limited industrial data, we adopt joint post-training with Supervised Fine-Tuning and Group Relative Policy Optimization using format, detection, and answer rewards. OmniAD achieves 79.9\% average accuracy on MMAD, outperforming strong open-source and proprietary MLLM baselines, and also delivers robust threshold-free performance on four industrial anomaly detection benchmarks. These results suggest that strengthening intrinsic perception within MLLMs is a practical route toward accurate and interpretable industrial anomaly reasoning.

\end{abstract}

\maketitle

\section{Introduction}
\label{sec:intro}
Anomaly detection is a core problem in industrial inspection because practical systems must not only flag defective products but also support diagnosis and downstream decision-making. Conventional anomaly detection methods typically output pixel-level or image-level predictions, which are insufficient for scenarios that require describing what is wrong, where it occurs, and why it matters. Recent Multimodal Large Language Models (MLLMs) have shown strong promise on open-domain visual reasoning~\cite{cao2023towards, zhu2024llms, li2023myriad}, but industrial anomaly analysis remains challenging because anomalies are often subtle, category-specific, and semantically entangled with fine-grained visual evidence.

A common strategy is to combine an MLLM with an anomaly detector as an external expert. The detector produces a heatmap or mask, and the MLLM uses this cue to answer anomaly-related questions. While intuitive, this expert-aid paradigm is less reliable than it appears and can even degrade anomaly understanding performance, as illustrated in Figure~\ref{fig:teaser}(Right). A key reason is that the detector and the MLLM are optimized with different objectives: the former emphasizes localization scores, whereas the latter requires semantically coherent evidence for reasoning. This mismatch makes the transferred cue useful in some cases but unstable across defect types and visual contexts. We attribute this failure to two issues. \textbf{1) Noisy visual cues.} External detector outputs are often threshold-dependent and imperfect; once converted into a hard cue, their errors are directly exposed to the MLLM. \textbf{2) Contextual suppression.} Even when the cue roughly localizes the anomaly, it can dominate the model's attention and weaken holistic reasoning about the object, surrounding structure, and failure mode. As a result, the MLLM may become better at ``looking where pointed'' but worse at conducting complete analysis.
We argue that these two issues share the same root cause: anomaly localization and anomaly understanding are loosely coupled through externally injected cues, rather than unified within a single reasoning process. In other words, the model receives anomaly evidence as an isolated hint, not as an internally generated and verifiable reasoning state, which limits both robustness and interpretability.

To address these limitations, we propose OmniAD, a framework that intrinsically integrates anomaly detection and understanding into a single multimodal reasoning process, as shown in Figure~\ref{fig:teaser}(Left). Instead of receiving anomaly cues from an external detector, OmniAD generates anomaly evidence within the same model and reuses it for downstream analysis within the same reasoning chain. Concretely, OmniAD first performs \emph{Visual Reasoning}, which localizes anomalies through Semantic Anomaly Encoding (SAE) and converts them into a compact textual intermediate representation. It then performs \emph{Visual Guided Textual Reasoning} (VGTR), where the internally generated anomaly evidence serves as a semantic anchor for high-level analysis. This intrinsic design improves cue reliability and preserves reasoning continuity, thereby mitigating both noisy visual cues and contextual suppression.

To cope with limited industrial training data, we further adopt a joint training strategy that combines supervised fine-tuning (SFT) and Group Relative Policy Optimization (GRPO). This combination improves both format compliance and task accuracy, leading to better generalization in both anomaly detection and understanding. Experimental results verify the effectiveness of OmniAD on both tasks. On MMAD~\cite{jiang2024mmad}, OmniAD achieves 79.9\% average accuracy, outperforming Qwen2.5-VL-7B~\cite{bai2025qwen2} by 15.9 points and GPT-4o by 5.0 points. On four anomaly detection benchmarks~\cite{bergmann2019mvtec, bergmann2022beyond, zou2022spot, zhang2024pku}, OmniAD delivers strong pixel-level and image-level performance without requiring threshold selection at inference time.

We summarize our contributions as follows:
\begin{itemize}
\item We propose OmniAD, a end-to-end framework that intrinsically integrates anomaly detection and understanding within a multimodal reasoning process, replacing brittle expert-aid coupling with internal anomaly evidence flow.

\item We develop a Multimodal Reasoning framework in which Semantic Anomaly Encoding generates anomaly evidence and Visual Guided Textual Reasoning reuses it for downstream analysis, forming an intrinsic generation-consumption chain.

\item We show that joint training with SFT and GRPO improves both anomaly detection and understanding, and extensive experiments on MMAD and four anomaly detection benchmarks validate the effectiveness of the design.
\end{itemize}

\section{Related Work}
\label{sec:RelatedWork}

\subsection{Industrial Anomaly Detection}
Traditional anomaly detection methods intend to discriminate and localize defects using only normal samples during training, and can generally be categorized into two types. Embedding-based methods~\cite{roth2022towards, hyun2024reconpatch, jiang2022softpatch, defard2021padim, guo2024dinomaly} typically extract embeddings from a pre-trained model and map them into a compact subspace or a memory bank for comparison to detect anomalies. Reconstruction-based methods~\cite{he2024mambaad, fan2024revitalizing, zavrtanik2021draem, zhang2024realnet, zhang2023diffusionad, fang2023fastrecon} train a reconstruction model using only normal samples. By failing to accurately reconstruct anomalous regions, reconstruction errors serve as anomaly scores.

Recently, few-shot anomaly detection methods have focused on modeling anomalies using a limited number of training samples. Recent approaches~\cite{jeong2023winclip, zhou2023anomalyclip, cao2024adaclip, chen2024clip, ma2025aa, gu2024anomalygpt, cao2023segment, gu2024univad} leverage foundation models, including CLIP~\cite{radford2021learning}, large language models, and SAM~\cite{kirillov2023segment}, for anomaly detection through prompt learning, multi-scale feature aggregation, semantic decoding, and patch-level feature matching.

However, many of these methods rely considerably on manually selected thresholds, which can reduce performance robustness. In contrast, our OmniAD treats segmentation as a text generation task, thereby avoiding reliance on such thresholds. This offers greater applicability across various anomaly detection tasks.

\subsection{Industrial Anomaly Understanding}
Recent years have witnessed significant advances in MLLMs for industrial anomaly understanding. Early studies combine expert models with MLLMs for anomaly localization and instruction~\cite{li2023myriad, gu2024anomalygpt}. Although textual output avoids thresholding for image-level detection, pixel-level detection still requires a threshold. Subsequent methods enhance logical anomaly detection and reasoning through text feature memory banks, executable code generation, customized lightweight multimodal models, and multi-stage training~\cite{jin2025logicad, zhang2024logicode, li2025lad}. Other approaches improve reasoning by integrating textual and visual knowledge, applying reinforcement learning, or multiple expert modules~\cite{jiang2024fabgpt, chao2025anomalyr1, chen2025can}. Meanwhile, MMAD~\cite{jiang2024mmad} provides a comprehensive benchmark for evaluating anomaly understanding.

Although MLLMs possess strong visual analysis and reasoning capabilities, it is still difficult to localize anomalies accurately and to seamlessly integrate their operations with the outputs of expert models. Our OmniAD addresses these current limitations with a new unified paradigm for detection and understanding.

\subsection{Visual Reinforcement Learning}
With the emergence of reasoning models such as o1~\cite{jaech2024openai} and DeepSeek-R1~\cite{guo2025deepseek}, visual reasoning has been introduced to better perform vision tasks. These reasoning capabilities primarily result from reinforcement learning, especially Group Relative Policy Optimization (GRPO)~\cite{shao2024deepseekmath} with verifiable reward. VLM-R1~\cite{shen2025vlm} first proposed a framework that showed improvement compared to SFT, especially on real-world benchmarks. Visual-RFT~\cite{liu2025visual} further extends efficient learning under limited data conditions, demonstrating strong generality. Seg-ZERO~\cite{liu2025seg} verified the generality of GRPO in segmentation tasks involving reasoning.

Some works have attempted to apply GRPO for anomaly understanding because its reasoning process can provide more interpretable information and exhibits strong generality under limited supervision. LAD-Reasoner~\cite{li2025lad} was proposed for logical anomaly detection, offering a readable reasoning process; however, it was not directly capable of anomaly understanding tasks. AnomalyR1~\cite{chao2025anomalyr1} designed a reasoned outcome alignment metric for an end-to-end framework for reasoning and localization, but it also treated the two tasks independently, and its bounding box format is not suitable for the variable shapes of anomaly regions. To fill this gap, our OmniAD is designed to unify detection and understanding, employing GRPO and enabling patch-level accurate localization.

\section{Method}

\label{sec:Overview}
We propose OmniAD, an end-to-end framework that unifies industrial anomaly detection and understanding within a single multimodal reasoning process. Rather than passing anomaly cues from an external detector to a downstream MLLM, OmniAD generates and consumes anomaly evidence within the same model and reasoning chain. This unified design improves cue reliability and preserves reasoning continuity, thereby addressing both noisy visual cues and contextual suppression. The entire framework is optimized through joint post-training with SFT and GRPO, enabling unified anomaly localization and understanding.

\subsection{Multimodal Reasoning}
\label{sec:Multimodal Reasoning}
\begin{figure}
    \centering
    \includegraphics[width=0.45\textwidth]{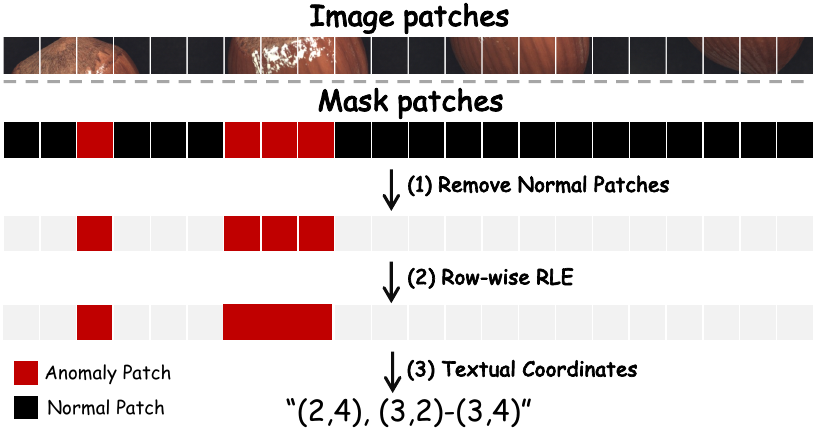}
    \caption{An illustration of the Semantic Anomaly Encoding process. White marked regions indicate anomalies. RLE denotes run-length encoding~\cite{golomb1966run}.}
    \Description{An illustration of the Semantic Anomaly Encoding process.}
    \label{fig:text-as-mask}
\vspace{-1.5em}
\end{figure}

We enhance fine-grained anomaly perception and analysis by integrating detection evidence directly into Multimodal Reasoning. Since expert-model assistance often degrades downstream analysis~\cite{jiang2024mmad}, our goal is not to provide stronger external cues, but to internalize anomaly evidence into the model's own reasoning process. In OmniAD, anomaly evidence is first generated through Visual Reasoning and then consumed by Visual Guided Textual Reasoning for downstream analysis. This unified design removes the external detector-to-MLLM handoff and turns localization into an internal intermediate variable of the reasoning chain.

\subsubsection{Visual Reasoning via Semantic Anomaly Encoding}
We propose Visual Reasoning to generate interpretable localization evidence for subsequent fine-grained analysis. Instead of importing thresholded heatmaps or masks from an external detector, Visual Reasoning can perform localization inside the MLLM itself and produce anomaly evidence as part of the reasoning process. Inspired by recent progress in text-based segmentation~\cite{lan2024text4seg}, Semantic Anomaly Encoding (SAE) serves as the representation that reformulates anomaly detection as text generation. Figure~\ref{fig:text-as-mask} illustrates the process. We first divide an image into fixed-size patches. A straightforward strategy would label every patch as either normal or anomalous, but anomaly masks are usually sparse and irregular, which would introduce a severe imbalance between positive and negative tokens. We therefore redesign the encoding for industrial anomalies. Specifically, we discard normal patches and retain only anomalous ones, following the one-class intuition widely used in anomaly detection. We then apply run-length encoding (RLE) along each row to compress contiguous anomaly patches, and convert the compressed sequence into a coordinate string. Compared with dense patch labeling, SAE is shorter, more information-dense, and easier for the language model to reuse in later reasoning steps. In this way, Visual Reasoning makes anomaly evidence intrinsic, threshold-free, and directly compatible with downstream reasoning.

\subsubsection{Visual Guided Textual Reasoning}
Building on Visual Reasoning, Visual Guided Textual Reasoning (VGTR) performs high-level reasoning conditioned on internally generated anomaly evidence. Instead of injecting an external cue that may dominate attention, VGTR reuses the evidence produced in the first stage within the same reasoning chain. This design directly addresses the loose coupling highlighted in the introduction: anomaly localization and anomaly understanding are no longer passed across model boundaries, but linked as consecutive states of one unified process. Consequently, the model receives anomaly evidence as a semantically coherent intermediate variable rather than an isolated hard hint, which improves robustness across defect types and visual contexts.

The anomaly type and location estimated by Visual Reasoning act as semantic anchors that guide the model toward relevant regions without overriding holistic analysis of object semantics, surrounding structure, and failure mode. We use a structured response format, \texttt{<seg>...</seg>}, \texttt{<reasoning>...</reasoning>}, and \texttt{<answer>...</answer>}, to explicitly separate localization, reasoning, and final decision. The \texttt{<seg>} sequence stores the SAE representation, \texttt{<reasoning>} contains the model's analysis of object state and defect evidence, and \texttt{<answer>} provides the task-specific conclusion. This structure establishes a transparent path from local anomaly evidence to global interpretation and makes the generation-consumption loop explicitly verifiable. As a result, VGTR mitigates both noisy cue propagation and contextual suppression in expert-aid pipelines, encouraging more balanced and complete full-image analysis.

\subsection{Joint Training}

\label{sec:Training}

To strengthen multimodal reasoning and inject industrial knowledge, we adopt an integrated training strategy that combines SFT and GRPO. SFT teaches the model the target response format, object-specific knowledge, and the coupling between localization and explanation. GRPO is then used to refine reasoning behavior using only verifiable rewards, without relying on expensive token-level reasoning annotations. In practice, we find that SFT provides the necessary task prior, while GRPO improves exploration and response quality. Their combination is substantially more effective than either stage alone.

\begin{figure}
    \centering
    \includegraphics[width=0.45\textwidth]{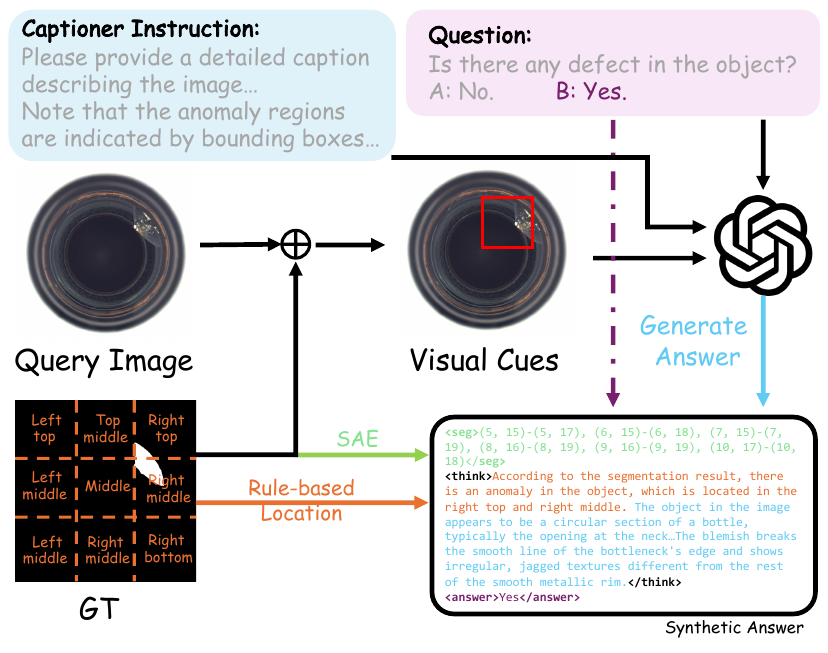}
    \vspace{-1em}
    \caption{An illustration of the data curation process.}
    \Description{An illustration of the data curation process.}
    \label{fig:data_process}
\vspace{-1.5em}
\end{figure}
\begin{table*}[ht]
\caption{The performance comparison of both proprietary and open-source MLLMs in MMAD is presented. All methods are evaluated in a 1-shot setting. Anomaly Discrimination uses the average accuracy across the normal and abnormal categories.} 
\vspace{-1em}
\centering
\setlength\tabcolsep{3pt}
\resizebox{\linewidth}{!}{
\begin{tabular}{lcccccccccccc}
\toprule
&    & Anomaly    && \multicolumn{4}{c}{Defect}  && \multicolumn{2}{c}{Object} &                              \\ \cmidrule{3-3} \cmidrule{5-8} \cmidrule{10-11}
\multirow{-2}{*}{Model}     & \multirow{-2}{*}{Scale} & Discrimination && Classification & Localization & Description & Analysis && Classification & Analysis && \multirow{-2}{*}{Average} \\ \midrule
Random Chance & - & 50.0 && 25.0 & 25.0 & 25.0 & 25.0 && 25.0 & 25.0 && 28.6 \\
\midrule
Human (expert) \cite{jiang2024mmad} & - & 95.2 && 75.0 & 92.3 & 83.3 & 94.2 && 86.1 & 80.4 & &86.7 \\
Human (ordinary) \cite{jiang2024mmad} & - & 86.9 && 66.3 & 85.6 & 71.3 & 81.5 && 89.6 & 69.7 && 78.7 \\
\midrule
Claude-3.5-sonnet & - & 60.1 && 60.1 & 48.8 & 67.1 & 79.1 && 85.2 & 79.8 && 68.4 \\
Gemini-1.5-flash \cite{team2024gemini} & - & 58.6 && 54.7 & 49.1 & 66.5 & 82.2 && 91.5 & 79.7 && 68.9 \\
Gemini-1.5-pro \cite{team2024gemini} & - & 68.6 && 60.1 & 58.6 & 70.4 & 82.5 && 89.2 & 82.3 && 73.1 \\
GPT-4o-mini & - & 64.3 && 48.6 & 38.8 & 63.7 & 80.4 && 88.6 & 79.7 && 66.3 \\
GPT-4o & - & 68.6 && 65.8 & 55.6 & 73.2 & 83.4 && 95.0 & 82.8 && 74.9 \\
\midrule
AnomalyGPT \cite{gu2024anomalygpt} & 7B & 65.6 && 27.5 & 28.0 & 36.9 & 32.1 && 29.8 & 35.8 && 36.5 \\
LLaVA-1.5 \cite{liu2024improved} & 7B & 51.3 && 37.0 & 36.6 & 50.6 & 69.8 && 68.3 & 69.5 && 54.7 \\
LLaVA-OneVision \cite{li2024llava} & 7B & 51.8 && 46.1 & 41.9 & 62.2 & 69.7 && 90.3 & 80.9 && 63.3 \\
MiniCPM-V2.6 \cite{yao2024minicpm} & 8B & 57.3 && 49.2 & 43.3 & 65.9 & 75.2 & &92.0 & 80.8 && 66.3 \\
InternVL2 \cite{chen2024internvl} & 8B & 60.0 && 43.9 & 47.9 & 57.6 & 78.1 && 74.2 & 80.4 && 63.1 \\
LLaVA-1.5 \cite{liu2024improved} & 13B & 50.0 && 38.8 & 46.2 & 58.2 & 73.1 && 73.6 & 71.0 && 58.7 \\
Qwen-2.5-VL \cite{bai2025qwen2} & 7B & 55.9 && 39.8 & 45.8 & 54.8 & 73.7 && 93.3 & 84.7 && 64.0 \\
LLaVA-NeXT \cite{liu2024llavanext} & 34B & 57.9 && 48.8 & 52.9 & 71.3 & 80.3 && 81.1 & 77.8 && 67.2 \\
InternVL2 \cite{chen2024internvl} & 76B & 68.3 && 54.2 & 56.7 & 66.3 & 80.5 && 86.4 & 82.9 && 70.8 \\
AnomalyR1 \cite{chao2025anomalyr1} & 3B & 60.2 && 63.5 & 70.1 & \textbf{80.4} & 85.2 && 82.4 & 86.1 && 76.9 \\
\midrule
\rowcolor{gray!20}
OmniAD(Ours) & 3B & {66.3} && {75.9} & {72.9} & 65.1 & 85.4 && 93.5 & 85.6 && 77.5 \\
\rowcolor{gray!20}
OmniAD(Ours) & 7B & \textbf{68.8} && \textbf{78.8} & \textbf{75.5} & 67.2 & \textbf{86.4} && \textbf{96.0} & \textbf{86.4} && \textbf{79.9} \\

\bottomrule
\end{tabular}
}

\label{tab: understanding}
\end{table*}

\begin{table*}[t]
\centering
\caption{Performance comparison of different expert visualization types and ground-truth (GT) guidance on MMAD. Values are reported as deltas relative to the baseline Qwen2.5-VL-7B~\cite{bai2025qwen2}; AnomalyCLIP~\cite{zhou2023anomalyclip} is used as the expert model.}
\vspace{-1em}
\setlength\tabcolsep{3pt}
\resizebox{\textwidth}{!}{%
    \begin{tabular}{c lccccccccccc}
    \toprule
    & & Anomaly && \multicolumn{4}{c}{Defect} && \multicolumn{2}{c}{Object} & \\
    \cmidrule{3-3} \cmidrule{5-8} \cmidrule{10-11}
    & \multirow{-2}{*}{Type} & Discrimination && Classification & Localization & Description & Analysis && Classification & Analysis && \multirow{-2}{*}{Average} \\ \midrule
     & - & 55.9 && 39.8 & 45.8 & 54.8 & 73.7 && 93.3 & 84.7 && 64.0 \\
    \hline
    \addlinespace[2pt]
    \multirow{5}{*}{\rotatebox[origin=c]{90}{Expert}} & bbox & \textcolor{blue}{-1.5} && \textcolor{red}{+2.2} & \textcolor{red}{+6.6} & \textcolor{red}{+2.3} & \textcolor{blue}{-2.5} && \textcolor{red}{+1.7} & \textcolor{blue}{-3.1} && \textcolor{red}{+0.8} \\
    & contour & \textcolor{blue}{-1.2} && \textcolor{blue}{-1.1} & \textcolor{red}{+6.1} & \textcolor{red}{+0.7} & \textcolor{blue}{-3.2} && \textcolor{red}{+1.6} & \textcolor{blue}{-3.2} && 0.0 \\
    & heatmap & \textcolor{blue}{-8.0} && \textcolor{blue}{-5.5} & \textcolor{red}{+7.6} & \textcolor{blue}{-2.7} & \textcolor{red}{+1.9} && \textcolor{blue}{-2.4} & \textcolor{blue}{-5.7} && \textcolor{blue}{-2.1} \\
    & highlight & \textcolor{blue}{-6.7} && \textcolor{blue}{-14.7} & \textcolor{red}{+6.5} & \textcolor{blue}{-11.0} & \textcolor{blue}{-3.3} && \textcolor{red}{+1.6} & \textcolor{blue}{-3.4} && \textcolor{blue}{-4.4} \\
    & mask & \textcolor{blue}{-6.0} && \textcolor{blue}{-2.4} & \textcolor{red}{+3.8} & \textcolor{blue}{-2.7} & \textcolor{red}{+0.1} && \textcolor{red}{+1.6} & \textcolor{blue}{-4.3} && \textcolor{blue}{-1.4} \\
    \hline
    \addlinespace[2pt]
    \multirow{5}{*}{\rotatebox[origin=c]{90}{GT}} & bbox & \textcolor{red}{+14.7} && \textcolor{red}{+8.5} & \textcolor{red}{+26.2} & \textcolor{red}{+9.4} & \textcolor{blue}{-1.1} && \textcolor{red}{+2.0} & \textcolor{blue}{-2.7} && \textcolor{red}{+8.2} \\
    & contour & \textcolor{red}{+16.4} && \textcolor{red}{+7.7} & \textcolor{red}{+23.9} & \textcolor{red}{+9.7} & \textcolor{blue}{-2.1} && \textcolor{red}{+1.8} & \textcolor{blue}{-2.4} && \textcolor{red}{+7.9} \\
    & highlight & \textcolor{red}{+17.3} && \textcolor{blue}{-3.6} & \textcolor{red}{+17.3} & \textcolor{red}{+0.3} & \textcolor{blue}{-2.3} && \textcolor{red}{+1.4} & \textcolor{blue}{-2.5} && \textcolor{red}{+4.0} \\
    & mask & \textcolor{red}{+9.8} && \textcolor{red}{+9.2} & \textcolor{red}{+5.7} & \textcolor{red}{+4.4} & \textcolor{blue}{-2.4} && \textcolor{red}{+1.1} & \textcolor{blue}{-2.4} && \textcolor{red}{+3.6} \\
    \midrule
    \rowcolor{gray!20}
    & \textbf{Ours} & \textcolor{red}{+5.5} && \textcolor{red}{+28.0} & \textcolor{red}{+27.7} & \textcolor{red}{+26.9} & \textcolor{red}{+13.8} && \textcolor{red}{+1.9} & \textcolor{red}{+2.2} && \textcolor{red}{+15.1} \\
    \bottomrule
    \end{tabular}%
}
\vspace{-0.5em}
\label{tab:expert_performance}
\end{table*}

After DeepSeekR1~\cite{guo2025deepseek} first introduced GRPO to enhance model reasoning capabilities, it has demonstrated potential across many tasks due to its adaptability. The core idea behind GRPO is the effectiveness and efficiency of verifiable rewards. GRPO is particularly well-suited for anomaly detection and analysis, as it often involves verifiable results, making relative optimization across multiple solutions a natural fit. In our task, the verifiable reward function plays a crucial role in improving model performance by providing clear and objective feedback during training. Therefore, we propose three reward functions to enhance unified anomaly detection and understanding capabilities.

\subsubsection{Multimodal Reasoning Format Reward}
This reward is designed to encourage the model to engage in a structured thinking process. It guides the model to output its visual reasoning within \verb'<seg>' and \verb'</seg>', its textual reasoning within \verb'<think>' and \verb'</think>', and the final answer between \verb'<answer>' and \verb'</answer>'. A reward of 1 is given if the format is followed; otherwise, the reward is 0.

\subsubsection{Detection Accuracy Reward}
In contrast to previous reward designs \cite{liu2025seg, liu2025visual} for segmentation tasks that evaluate predictions using Intersection-over-Union (IoU), we use the F1-score as the evaluation metric because it better focuses on precision and recall in class-imbalanced conditions. The process first converts the SAE format prediction into a mask, and then computes the F1-score between the predicted mask and the ground truth. The reward is defined as:
\vspace{-0.5em}

\begin{equation}
R_{\text{Acc}} =
\begin{cases}
1 & \text{if } \mathcal{F}_{\mathcal{G}} = \emptyset \text{ and } \mathcal{F}_{\mathcal{P}} = \emptyset \\
0 & \text{if } \mathcal{F}_{\mathcal{G}} = \emptyset \text{ and } \mathcal{F}_{\mathcal{P}} \neq \emptyset \\
\alpha \cdot \text{F1-score} & \text{if } \mathcal{F}_{\mathcal{G}} \neq \emptyset
\end{cases}
\end{equation}
\vspace{-0.5em}

where $\mathcal{F}_{\mathcal{G}}$ and $\mathcal{F}_{\mathcal{P}}$ represent the anomaly regions in the ground truth and prediction, respectively. $\alpha$ is a scale factor used to balance with other rewards, as the F1-score is typically a relatively small number. This reward function incentivizes higher rewards for more accurate detections and provides no reward for detections in normal images.

\subsubsection{Answer Accuracy Reward}
This reward evaluates the final answer against the ground truth and directly encourages correct task completion. A reward of 1 is given for a correct answer, while a reward of 0.1 is assigned for an incorrect answer. Compared with a strict zero-reward design, this formulation provides a smoother optimization signal during RL training and helps stabilize learning when the model explores different reasoning paths.

\subsection{Data Curation}
To train OmniAD for unified detection and understanding, we construct a multimodal dataset tailored to both stages. SFT requires dense supervision for reasoning and final answers, but such data is scarce in industrial anomaly analysis because annotation demands domain knowledge. We therefore build a data curation pipeline for high-quality VQA, as illustrated in Figure~\ref{fig:data_process}. We sample one example per category from several industrial anomaly datasets~\cite{yang20253cad, zhang2023industrial, arodi2024cableinspect, yang2024defect, huang2020surface, wang2024real}, and follow~\cite{chao2025anomalyr1} to improve cross-dataset coverage and reduce domain gaps. For GRPO, dense reasoning annotations are unnecessary, so we reuse the same data sources with answer-level supervision and verifiable localization targets. The final training set contains 1.6K images with 6.4K QA pairs for SFT and 293 images with 1.1K QA pairs for GRPO. For one-shot inference, we prepend the prompt with \verb'<image> This is an image of a normal object.', where \verb'<image>' denotes a randomly selected normal reference image from the same category as the query.

\section{Experiments}

\begin{table*}
\caption{\textbf{Comparison of pixel-level anomaly detection performance in both 0-shot and 1-shot settings.} Detection results are report as (F1-score, Acc). }
\vspace{-1em}
\renewcommand{\arraystretch}{1.4} 
\setlength{\tabcolsep}{3pt}      
\centering
\resizebox{\linewidth}{!}{
\begin{tabular}{l cc cc cc cc cc} 
\toprule 
\multirow{2}{*}{Methods} & \multicolumn{2}{c}{MVTec-AD}                 & \multicolumn{2}{c}{VisA}                    & \multicolumn{2}{c}{MVTec-LOCO}               & \multicolumn{2}{c}{GoodsAD}                  & \multicolumn{2}{c}{Average}                    \\ \cline{2-11} 
                         & \multicolumn{1}{c}{0-shot}      & 1-shot      & \multicolumn{1}{c}{0-shot}      & 1-shot      & \multicolumn{1}{c}{0-shot}      & 1-shot      & \multicolumn{1}{c}{0-shot}      & 1-shot      & \multicolumn{1}{c}{0-shot}      & 1-shot      \\ \hline 
WinCLIP~\cite{jeong2023winclip}                  & \multicolumn{1}{c}{19.0, 90.7} & \multicolumn{1}{c}{19.8, 90.1} & \multicolumn{1}{c}{8.4, 98.1}  & \multicolumn{1}{c}{12.5, {98.8}} & \multicolumn{1}{c}{10.4, 18.5} & \multicolumn{1}{c}{10.4, 85.2} & \multicolumn{1}{c}{10.6, 94.4} & \multicolumn{1}{c}{10.1, 88.7} & \multicolumn{1}{c}{12.1, 75.4 }                  & \multicolumn{1}{c}{13.2, 90.7 }                   \\
AnomalyGPT~\cite{gu2024anomalygpt}               & \multicolumn{1}{c}{23.4, 92.9} & \multicolumn{1}{c}{{42.7}, {96.1}} & \multicolumn{1}{c}{10.8, 98.2} & \multicolumn{1}{c}{\textbf{32.0}, {98.8}} & \multicolumn{1}{c}{13.6, {81.8}} & \multicolumn{1}{c}{{24.7}, 86.5}  & \multicolumn{1}{c}{7.2, 88.9}  & \multicolumn{1}{c}{11.2, 79.1} & \multicolumn{1}{c}{13.8, 90.5}                  &\multicolumn{1}{c}{27.7, 90.1}                   \\
AnomalyCLIP~\cite{zhou2023anomalyclip}              & \multicolumn{1}{c}{34.1, {95.9}}         & -           & \multicolumn{1}{c}{17.8, 97.8}         & -           & \multicolumn{1}{c}{\textbf{15.9}, 61.1}         & -           & \multicolumn{1}{c}{{16.8}, \textbf{96.9}} & -           & \multicolumn{1}{c}{{21.2}, 71.1}                  & -           \\
AdaCLIP~\cite{cao2024adaclip}                  & \multicolumn{1}{c}{30.2, 95.5}         &    -        & \multicolumn{1}{c}{\textbf{32.0}, \textbf{99.3}}         &    -         & \multicolumn{1}{c}{11.3, 6.0}         &    -        & \multicolumn{1}{c}{7.9, 94.8}         &    -        & \multicolumn{1}{c}{20.3, 73.9}                  &    -         \\
AA-CLIP~\cite{ma2025aa}                  & \multicolumn{1}{c}{{35.4}, 95.4} & -           & \multicolumn{1}{c}{4.7, 98.3}  & -           & \multicolumn{1}{c}{{15.6}, 79.9} & -           & \multicolumn{1}{c}{12.5, 94.6} & -           & \multicolumn{1}{c}{17.1, {92.1}}                  & -           \\
UniVAD~\cite{gu2024univad}                   & \multicolumn{1}{c}{-}           & \multicolumn{1}{c}{{42.7}, 96.0} & \multicolumn{1}{c}{-}           & \multicolumn{1}{c}{27.2, \textbf{99.1}} & \multicolumn{1}{c}{-}           & \multicolumn{1}{c}{\textbf{28.1}, {88.4}} & \multicolumn{1}{c}{-}           & \multicolumn{1}{c}{{14.2}, {94.1}} & \multicolumn{1}{c}{-}            &    \multicolumn{1}{c}{\textbf{28.1}, {94.4}}                   \\
\rowcolor{gray!20}
\textbf{OmniAD}          & \multicolumn{1}{c}{\textbf{45.6}, \textbf{96.9}} & \multicolumn{1}{c}{\textbf{52.1}, \textbf{97.6}} & \multicolumn{1}{c}{{18.6}, {98.6}} & \multicolumn{1}{c}{{29.5}, \textbf{99.1}} & \multicolumn{1}{c}{14.4, \textbf{89.9}} & \multicolumn{1}{c}{14.7, \textbf{91.7}} & \multicolumn{1}{c}{\textbf{20.4}, {96.4}} & \multicolumn{1}{c}{\textbf{26.7}, \textbf{97.4}} & \multicolumn{1}{c}{\textbf{23.3}, \textbf{95.9}} & \multicolumn{1}{c}{{27.9}, \textbf{96.8}} \\
\bottomrule 
\end{tabular}
}

\label{tab:pixel_level}
\end{table*}
\begin{table*}[ht]
\caption{\textbf{Comparison of image-level anomaly detection performance in both 0-shot and 1-shot settings.} Detection results are report as (F1-score, Acc).}
\vspace{-1em}
\renewcommand{\arraystretch}{1.4} 
\setlength{\tabcolsep}{3pt}      
\centering
\resizebox{\linewidth}{!}{
\begin{tabular}{l cc cc cc cc cc} 
\toprule
\multirow{2}{*}{Methods} & \multicolumn{2}{c}{MVTec-AD}             & \multicolumn{2}{c}{VisA}                 & \multicolumn{2}{c}{MVTec-LOCO}           & \multicolumn{2}{c}{GoodsAD}              & \multicolumn{2}{c}{Average}              \\ \cline{2-11}
                         & \multicolumn{1}{c}{0-shot}    & 1-shot    & \multicolumn{1}{c}{0-shot}    & 1-shot    & \multicolumn{1}{c}{0-shot}    & 1-shot    & \multicolumn{1}{c}{0-shot}    & 1-shot    & \multicolumn{1}{c}{0-shot}    & 1-shot    \\ \hline 
WinCLIP~\cite{jeong2023winclip}                  & \multicolumn{1}{c}{74.5, 65.8} & 70.1, 62.7 & \multicolumn{1}{c}{64.7, 62.2} & 57.4, 61.3 & \multicolumn{1}{c}{{77.3}, {63.2}} & 75.5, 62.0 & \multicolumn{1}{c}{57.9, 53.0} & 68.3, 55.2 & \multicolumn{1}{c}{68.6, 61.0}          & \multicolumn{1}{c}{67.8, 60.3}          \\
AnomalyGPT~\cite{gu2024anomalygpt}               & \multicolumn{1}{c}{83.2, 72.8} & {89.8}, {84.9} & \multicolumn{1}{c}{66.6, 56.5} & {81.9}, {77.5} & \multicolumn{1}{c}{77.2, 63.0} & 77.3, 63.2 & \multicolumn{1}{c}{71.1, 55.3} & 71.1, 55.4 & \multicolumn{1}{c}{74.5, 61.9}          & \multicolumn{1}{c}{{80.0}, {70.2}} \\
AnomalyCLIP~\cite{zhou2023anomalyclip}             & \multicolumn{1}{c}{{86.7}, {79.9}} & -         & \multicolumn{1}{c}{75.4, 64.5} & -         & \multicolumn{1}{c}{{77.3}, {63.2}} & -         & \multicolumn{1}{c}{71.1, 55.4} & -         & \multicolumn{1}{c}{{77.6}, 65.7} & -          \\ 
AdaCLIP~\cite{cao2024adaclip}                  & \multicolumn{1}{c}{81.9, 73.8} & -         & \multicolumn{1}{c}{{78.7}, {75.5}} & -         & \multicolumn{1}{c}{{77.3}, {63.2}} & -         & \multicolumn{1}{c}{71.1, 56.2} & -         & \multicolumn{1}{c}{77.2, {67.1}}                    &            \\ 
AA-CLIP~\cite{ma2025aa}                  & \multicolumn{1}{c}{85.2, 77.6} & -         & \multicolumn{1}{c}{66.4, 54.2} & -         & \multicolumn{1}{c}{77.2, 63.1} & -         & \multicolumn{1}{c}{{71.5}, {56.5}} & -         & \multicolumn{1}{c}{75.0, 62.8}          & -          \\
UniVAD~\cite{gu2024univad}                   & \multicolumn{1}{c}{-}         & 88.5, 82.7 & \multicolumn{1}{c}{-}         & 79.4, 75.5 & \multicolumn{1}{c}{-}         & {77.6}, {64.1} & \multicolumn{1}{c}{-}         & {71.3}, {56.5} & \multicolumn{1}{c}{-}                   & \multicolumn{1}{c}{79.2, 69.7}          \\
\rowcolor{gray!20}
\textbf{OmniAD}          & \multicolumn{1}{c}{\textbf{97.6}, \textbf{96.5}} & \textbf{97.2}, \textbf{96.0} & \multicolumn{1}{c}{\textbf{89.2}, \textbf{87.4}} & \textbf{87.5}, \textbf{86.6} & \multicolumn{1}{c}{\textbf{92.6}, \textbf{90.3}} & \textbf{93.2}, \textbf{91.1} & \multicolumn{1}{c}{\textbf{90.0}, \textbf{87.9}} & \textbf{90.1}, \textbf{88.4} & \multicolumn{1}{c}{\textbf{92.2}, \textbf{90.1}} & \multicolumn{1}{c}{\textbf{92.0}, \textbf{91.1}} \\ \bottomrule
\end{tabular}
}
\vspace{-1em}
\label{tab:image_level}
\end{table*}
\subsection{Experimental Settings}
\subsubsection{Implementation Details}
OmniAD is built upon Qwen2.5-VL~\cite{bai2025qwen2}, an open-source MLLM chosen for its strong balance between performance and efficiency. All images are resized to $512 \times 512$ during both training and inference to mitigate the impact of varying input resolutions. The Semantic Anomaly Encoding is configured with a patch grid of $24 \times 24$, sufficient to capture fine-grained anomaly patterns while maintaining computational efficiency. The scale factor $\alpha$ in the Detection Accuracy Reward is empirically set to 2 to balance contribution with other reward objectives. Training is conducted on 4$\times$H800 GPUs using DeepSpeed~\cite{rasley2020deepspeed} for distributed optimization, with parameter-efficient fine-tuning via LoRA~\cite{hu2021lora} rank 4 and a total batch size of 16. For the supervised fine-tuning (SFT) stage, we use an initial learning rate of $1 \times 10^{-4}$ and weight decay of 0.03 to establish a stable task prior. For the subsequent GRPO refinement stage, the initial learning rate is reduced to $1 \times 10^{-6}$ with increased weight decay of 0.1, and 16 rollouts are generated at each training step. 
These settings reflect SFT learns robust priors, while GRPO stabilizes reasoning refinement with smaller updates and diverse rollouts.

\subsubsection{Benchmarks} 
As OmniAD is designed as a unified method for both anomaly detection and anomaly understanding, we evaluate it comprehensively on both tasks. For anomaly understanding, we adopt MMAD~\cite{jiang2024mmad}, which assesses seven subtasks through multiple-choice questions: Anomaly Discrimination, Defect Classification, Defect Localization, Defect Description, Defect Analysis, Object Classification, and Object Analysis. For anomaly detection, we evaluate on four representative benchmarks: MVTec-AD~\cite{bergmann2019mvtec}, VisA~\cite{zou2022spot}, MVTec-LOCO~\cite{bergmann2022beyond}, and GoodsAD~\cite{zhang2024pku}. MVTec-AD and VisA are widely used datasets that primarily emphasize textural anomalies, whereas MVTec-LOCO and GoodsAD place stronger emphasis on structural and logical defects. This benchmark combination covers diverse anomaly patterns and reasoning demands, providing a rigorous testbed for the proposed unified paradigm.

\subsubsection{Evaluation Metrics}
For anomaly detection, we report F1-score and accuracy at both pixel and image levels, following common practice for fair comparison with prior methods. Following Anomalib~\cite{akcay2022anomalib}, baseline methods use a dataset-level threshold selected to maximize the harmonic mean of image-level and pixel-level performance, and both levels are evaluated under this shared threshold. In contrast, OmniAD directly predicts anomaly regions in SAE format and therefore does not require threshold tuning at inference time, which better matches practical deployment settings. For anomaly understanding, we report multiple-choice accuracy on MMAD, which directly reflects end-task decision quality across all seven subtasks.

\subsubsection{Inference Setting}
For all experiments, the few-shot inference settings follow previous works~\cite{gu2024anomalygpt, jeong2023winclip}. A 0-shot setting means only a query image is provided, while a 1-shot setting includes an additional normal image from the same category as the query image for reference.

\begin{figure*}
    \centering
    \includegraphics[width=1\textwidth]{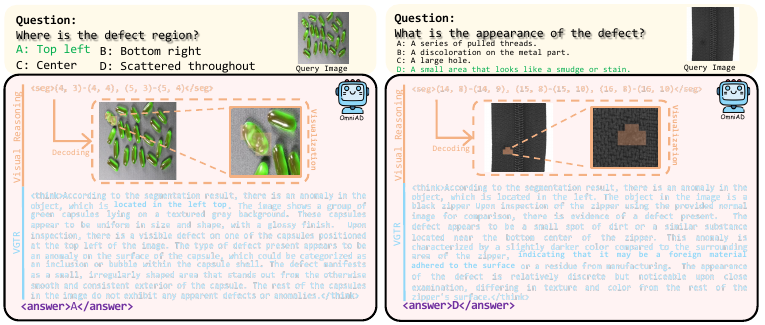}
    \vspace{-2em}
    \caption{Qualitative results on MMAD~\cite{jiang2024mmad}. The Multimodal Reasoning process facilitates accurate anomaly detection and question analysis. The GT answer is indicated by green.}
    \Description{Qualitative results on MMAD~\cite{jiang2024mmad}. The Multimodal Reasoning process facilitates accurate anomaly detection and question analysis. The correct answer is indicated by green.}
    \label{fig:vis}
    \vspace{-1em}
\end{figure*}

\subsection{Main Results}

\subsubsection{Anomaly Understanding}
Table~\ref{tab: understanding} compares OmniAD with proprietary and open-source MLLMs on MMAD. OmniAD achieves the highest overall accuracy, reaching 79.9\% with the 7B model. Its largest gains over Qwen2.5-VL-7B occur in Defect Classification, Defect Localization, and Defect Analysis, showing that domain-aligned anomaly perception is the main source of improvement and can outweigh the broader knowledge of GPT-4o. OmniAD also outperforms domain-specific baselines such as AnomalyGPT and AnomalyR1 on localization- and reasoning-intensive subtasks. Although AnomalyR1 remains competitive in Defect Description, these results suggest that coupling intrinsic localization with explanation is more effective than relying on an external detector or improving reasoning alone. OmniAD performs comparably to ordinary human annotators on average and surpasses them in several subtasks, including Defect Analysis and Object Classification. Figure~\ref{fig:vis} further demonstrates that the predicted SAE tokens provide meaningful evidence for downstream explanations.

Table~\ref{tab:expert_performance} directly evaluates the expert-aid paradigm using visual cues derived from an external anomaly detector or ground truth (GT). External cues provide limited improvements, mainly in Defect Localization, but generally reduce overall MMAD accuracy and often harm Anomaly Discrimination, Defect Classification, and Object Analysis. This suggests that noisy or overly restrictive expert cues can disrupt holistic reasoning. In contrast, GT guidance improves the average score, indicating that localization is beneficial when the cues are reliable and compatible with downstream reasoning. By generating and using anomaly evidence within the same reasoning chain, OmniAD avoids this trade-off and achieves stronger overall performance than both expert-aid and GT-style prompting.

\begin{figure}
\hspace{-0.1cm}
    \vspace{-1em}
	\centering
	\includegraphics[width=0.48\textwidth]{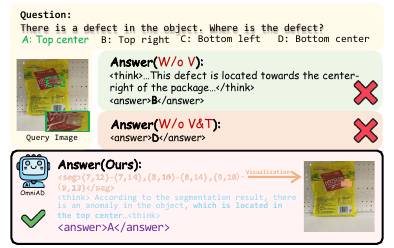}
    \vspace{-1em}
    \captionsetup{font=small}
	\caption{Ablation case study: qualitative comparison showing how removing Visual Reasoning weakens anomaly localization, and removing VGTR weakens reasoning quality.}
    \Description{Case Study of Multimodal Reasoning.}
    \label{fig:vis_abl_multimodal}
    \vspace{-1em}
\end{figure}  

\subsubsection{Anomaly Detection}
To evaluate whether multimodal reasoning improves detection quality, we compare OmniAD with six baselines under 0-shot and 1-shot settings. As shown in Table~\ref{tab:pixel_level}, OmniAD achieves the best average 0-shot pixel-level performance and remains competitive in the 1-shot setting without relying on thresholded heatmaps. Table~\ref{tab:image_level} shows that it achieves the best average image-level performance in both settings. Its consistent performance indicates that it does not depend on reference images but benefits from intrinsic anomaly reasoning. The gains are particularly notable on MVTec-LOCO and GoodsAD, which require reasoning about structures, spatial arrangements, and object-level consistency rather than only local texture deviations. This gives OmniAD an advantage over feature-matching methods dominated by local appearance similarity. Overall, the pixel-level results demonstrate that SAE provides precise local evidence, while the stronger image-level results highlight the effectiveness of unified reasoning for robust global judgments, confirming improvements in both anomaly perception and explanation.

\subsection{Ablation Study}

\subsubsection{Training Strategy}
We first compare different training strategies in Table~\ref{tab:abl_train}. Joint SFT+GRPO consistently outperforms either component alone on both anomaly understanding and detection, confirming that the two stages are complementary rather than interchangeable. In particular, GRPO-only training performs substantially worse, indicating that RL is effective for behavior refinement but insufficient for acquiring the domain knowledge and response structure required by industrial anomaly analysis. This observation is consistent with recent findings~\cite{su2025expanding} that RL mainly improves the efficiency of reaching correct behaviors once a strong prior has been established.
These results clarify the functional division between the two stages. SFT builds a stable prior over output format, object-level knowledge, and the coupling between localization and explanation; GRPO then improves result-level alignment under verifiable rewards. When GRPO is used alone, the model must simultaneously discover output structure, anomaly knowledge, and reasoning trajectories, which makes optimization unstable and sample-inefficient. Overall, the ablation supports using RL as a refinement stage on top of supervised adaptation, rather than as its replacement.

\begin{table}[t]
\centering
\caption{\textbf{Ablations on training strategy.} Detection results are reported as (F1-score, Acc).}
\vspace{-1em}
\small
\renewcommand{\arraystretch}{1.15}
\setlength{\tabcolsep}{3pt}
\resizebox{\linewidth}{!}{
\begin{tabular}{l ccccc c cc}
\toprule
 \multirow{2}{*}{Method} & \multicolumn{5}{c}{Understanding} && \multicolumn{2}{c}{Detection} \\ \cline{2-6} \cline{8-9}
 & Def. Loc. & Def. Ana. & Def. Des. & Obj. Ana. & Avg. && Pixel-level & Image-level \\ \hline
 SFT &64.6&77.7&69.1&76.6&69.8&&15.0, 93.4& 86.4, 81.7 \\ 
 GRPO &47.6&64.8&48.9&71.2&55.7&&18.8, \textbf{97.0}& 88.3, 85.9 \\
 \rowcolor{gray!20}
 \textbf{J.T.} &\textbf{73.5}&\textbf{87.5}&\textbf{81.7}&\textbf{86.9}&\textbf{79.1}&& \textbf{23.3}, 95.9&\textbf{92.2}, \textbf{90.1}\\
 \bottomrule
\end{tabular}
}
\label{tab:abl_train}
\vspace{-0.5em}
\end{table}

\begin{table}[t]
\centering
\caption{\textbf{Ablation on Multimodal Reasoning.} Def. Loc. and Obj. Ana. denote Defect Localization and Object Analysis. V and T denote Visual reasoning and Textual reasoning, respectively.}
\vspace{-1em}
\small
\renewcommand{\arraystretch}{1.15}
\begin{tabular}{l ccc}
\toprule
\multirow{2}{*}{Method} & \multicolumn{3}{c}{Understanding} \\
\cline{2-4}
& Def. Loc. & Obj. Ana. & Avg. \\ \hline
w/o V\&T &64.9 &72.2 &68.0 \\
w/o V &71.2 &83.6 &77.4 \\
w/o T &66.4 &75.6 &72.5 \\
\rowcolor{gray!20}
\textbf{All} &\textbf{73.5} &\textbf{86.9} &\textbf{79.1} \\
\bottomrule
\end{tabular}
\label{tab:abl_rea}
\vspace{-0.5em}
\end{table}

\subsubsection{Multimodal Reasoning}
We next analyze the contribution of Visual Reasoning and VGTR in Table~\ref{tab:abl_rea}. Combining both components yields the strongest performance by a clear margin. VGTR provides the high-level reasoning scaffold, while Visual Reasoning supplies precise and reusable anomaly evidence for downstream analysis. The full model outperforming either component alone supports our diagnosis of contextual suppression: internally generated cues can guide reasoning effectively, whereas weakly integrated cues cannot. Figure~\ref{fig:vis_abl_multimodal} shows a representative qualitative case.
The ablation further indicates that neither localization-oriented reasoning nor textual reasoning is sufficient in isolation. Removing Visual Reasoning causes a substantial drop because the model loses structured anomaly evidence, while removing VGTR also degrades performance because localization without semantic interpretation cannot complete anomaly understanding. Their gains are therefore complementary, not merely additive. This finding is consistent with our central claim that anomaly evidence should be generated and consumed within one continuous reasoning chain.

\subsubsection{Reward Function}
Table~\ref{tab:abl_reward} analyzes the reward design. Both rewards are beneficial, but the Answer Accuracy Reward has the larger overall effect, improving not only understanding but also detection. This suggests that optimizing the final decision in a structured reasoning pipeline also improves upstream perception quality through end-to-end credit assignment. The result further supports our design principle that perception and reasoning should be optimized jointly rather than as disconnected objectives.
At the same time, the Detection Accuracy Reward remains essential because it directly constrains SAE quality and discourages answer-level shortcut behaviors. In other words, the Answer Accuracy Reward promotes global consistency of reasoning and decisions, while the Detection Accuracy Reward stabilizes the intermediate visual evidence that later reasoning depends on. The best performance is obtained only when both rewards are used together, indicating that reliable anomaly understanding requires explicit supervision at both intermediate localization and final decision levels.
\begin{table}[t]
\centering
\caption{\textbf{Ablation on reward design.} AAR and DAR denote Answer Accuracy Reward and Detection Accuracy Reward. Detection results are reported as (F1-score, Acc).}
\small
\vspace{-1em}
\renewcommand{\arraystretch}{1.15}
\setlength{\tabcolsep}{5pt}
\resizebox{\linewidth}{!}{
\begin{tabular}{l cccc}
\toprule
\multirow{2}{*}{Reward} & Understanding && \multicolumn{2}{c}{Detection} \\
\cline{2-2} \cline{4-5}
& Avg. && Pixel-level & Image-level \\
\hline
w/o AAR & 70.7 && 17.4, 94.0 & 92.3, 90.2 \\
w/o DAR & 78.3 && 21.4, 95.6 & \textbf{92.4}, \textbf{90.5} \\
\rowcolor{gray!20}
\textbf{All} & \textbf{79.1} && \textbf{23.3}, \textbf{95.9} & 92.2, 90.1 \\
\bottomrule
\end{tabular}
}
\label{tab:abl_reward}
\vspace{-1.5em}
\end{table}

\section{Conclusion}

We presented OmniAD, a unified framework for industrial anomaly detection and understanding. Motivated by the limitations of the expert-aid paradigm, where external anomaly cues can be noisy and may suppress holistic reasoning, OmniAD integrates anomaly localization and understanding within a single multimodal reasoning chain. Specifically, it first uses Semantic Anomaly Encoding to generate compact textual anomaly evidence and then performs Visual Guided Textual Reasoning for downstream analysis. This design enables anomaly evidence to be generated and consumed within the same model, improving cue reliability while preserving the continuity of reasoning. With joint SFT and GRPO training, OmniAD achieves strong performance on both anomaly detection and understanding benchmarks. Extensive experiments further show that the proposed unified design is effective across different anomaly types and evaluation settings. Overall, these results suggest that, for industrial anomaly analysis, strengthening intrinsic anomaly perception and maintaining reasoning continuity within the MLLM is more effective than loosely coupling external detector outputs to downstream understanding.

\bibliographystyle{ACM-Reference-Format}
\bibliography{samples/sample-base}










\end{document}